\documentclass{cas-dc}
\usepackage{float}
\usepackage[numbers]{natbib}
\usepackage{subfigure} 
\def\tsc#1{\csdef{#1}{\textsc{\lowercase{#1}}\xspace}}
\tsc{WGM}
\tsc{QE}
\tsc{EP}
\tsc{PMS}
\tsc{BEC}
\tsc{DE}

\begin{document}
\let\printorcid\relax
\let\WriteBookmarks\relax
\def\floatpagepagefraction{1}
\def\textpagefraction{.001}
\shorttitle{}
\shortauthors{HeChen Yang et~al.}

\title[mode = title]{A Comparison for Patch-level Classification of Deep Learning Methods on Transparent Environmental Microorganism Images: from Convolutional Neural Networks to Visual Transformers
}  

\author[a]{Hechen Yang}[type=editor,style=chinese,auid=000,bioid=1,]
\author[a]{Chen Li}[style=chinese]
\cormark[1]
\ead{lichen201096@hotmail.com}
\author[a]{Jinghua Zhang}[style=chinese]
\author[a]{Peng Zhao}[style=chinese]
\author[a]{Ao Chen}[style=chinese]
\author[b]{Xin Zhao}[style=chinese]
\author[c]{Tao Jiang}[style=chinese]
\author[d]{Marcin Grzegorzek}

\address[a]{Microscopic Image and Medical Image Analysis Group, MBIE College, Northeastern University, 110169, Shenyang, PR China}
\address[b]{Environmental Engineering Department, Northeastern University, Shenyang 110169, China}
\address[c]{School of Control Engineering, Chengdu 
University of Information Technology,Chengdu 610225, China}
\address[d]{Institute of Medical Informatics, University of Luebeck, Luebeck, Germany}

\begin{abstract}
Nowadays, analysis of \emph{Transparent Environmental Microorganism Images} (T-EM images) in the field of 
computer vision has gradually become a new and interesting spot. 
This paper compares different deep learning classification 
performance for the problem that T-EM images are 
challenging to analyze.
We crop the T-EM images into 8×8 and 224×224 pixel patches in the same 
proportion and then divide the two different pixel patches 
into foreground and background according to ground truth. 
We also use four convolutional neural networks and a 
novel ViT network model to compare the foreground and 
background classification experiments. 
We conclude that ViT performs the worst in classifying 
$8 \times 8$ pixel patches, but it outperforms most 
convolutional neural networks in classifying $224 \times 224$ pixel patches. 
\end{abstract}


\maketitle

\section{Introduction}
Nowadays in industrialized countries, environmental pollution is an essential problem, and microorganisms are closely related to environmental protection and sustainable development. \emph{Environmental Microorganisms} (EMs) are tiny living beings in natural or artificial environments, which are invisible to the naked eye (their size usually varies between 0.1 and 100 µm) and they can only be observed under microscopes. EMs are natural decomposers and indicators and they can solve environmental pollution without any secondary pollution. For example, \emph{Actinophrys} can digest organic waste in sludge and improve the quality of fresh water. Therefore, EMs plays an important role in solving the problem of environmental pollution.

There are four traditional methods for EM identification. The first one is physical method. This method has high accurate in identifying EMs but requires expensive equipment. The second is chemical method, which has a high accuracy rate but often prone to cause secondary pollution. The third method is molecular biology recognizes EMs through the sequence of genes. However, it requires professional researchers and expensive equipment. The fourth method is morphological observation, which  requires experienced operators to observe the EMs under a microscope and it requires a lot of manpower and energy. Therefore, these four methods have their shortcomings in actual operation. Therefore, we consider that the excellent performance of deep learning in the field of image processing can be used to make up for the shortcomings of traditional methods. 

In EMs, there are many categories that are transparent and colorless, resulting a lot of difficulties in the EM image analysis process.  
Nowadays, the application of \emph{Transparent Environmental Microorganism Images} (T-EM images) has become more and more widely used in various fields around humans. 
Such as, 
identifying the number of EMs in the environment to judge the 
degree of environmental pollution~\cite{li-2021-EMIDF}. In recent 
years, the detection of transparent objects in images is also 
a hot spot in vision research. It is not an easy task to 
detect  
whether there are transparent objects or translucent objects 
in images~\cite{khaing-2018-TODUC}. Because the transparent 
target area to be observed is generally very small or very 
thin, the colors and contrast of foreground and background 
are similar, and only the residual edge part leads to the low 
resolution of foreground or background, which largely depends 
on its background and lighting conditions. Therefore, there 
is an urgent need for some effective methods to identify 
transparent or translucent EM images.

In recent years, computer vision has good performance in 
computer vision acquisition~\cite{tenenbaum-1970-AICV}, 
contour tracking~\cite{isard-1996-CTBSP}, edge detection
~\cite{kelly-1970-EDIPB}, face recognition~\cite{bruce-1986-UFR}, fingerprint recognition~\cite{baldi-1993-NNFFR}, 
automatic driving~\cite{grigorescu-2020-ASODL}, and medical 
image analysis~\cite{shen-2017-DLIMI}. We considering the 
excellent performance of computer vision in image analysis, 
such as high speed, high accuracy, low consumption, 
high degree of quantification, strong objectivity
~\cite{jahne-2000-CVAA}, therefore computer vision can make up 
the 
shortcomings of traditional morphological methods. It brings 
new opportunities to EM images analysis. Especially 
when an image is transparent and short of visual information, 
we usually need to crop it into patches to discover more 
visual details to recover the lost information. Hence, 
research work on patch-level is significant for transparent image analysis, such as patch-level image segmentation and classification tasks.

In recent years, deep learning is the most efficient method 
in the field of machine vision, such as the popular 
\emph{Convolutional Neural Network} (CNN) Xception~\cite{carreira-1998-XATFT}, VGG-16~\cite{guan-2019-DCNNV}, 
Resnet50~\cite{reddy-2019-TLWRF}, Inception-V3~\cite{xia-2017-IVFFC}, MobileNet~\cite{chen-2018-AEHM}, 
NasNet~\cite{martinez-2020-PEOFT}, and novel 
{\emph{Visual Transformers} (VTs)~\cite{vaswani-2017-AIAYN}}. 
CNNs slowly expand the receptive field until it covers the 
whole image by accumulating convolution layers, so CNNs 
complete the extraction of graphics from local to global 
information. In contrast, transformers can obtain global 
information from the beginning, 
so they are more difficult to learn, but their ability to 
learn long-term dependence is stronger~\cite{vaswani-2017-AIAYN}. Hence, CNNs and Transformers have advantages and 
disadvantages in dealing with visual information. 
Therefore, this paper compares the  patch-level classification 
performance of transparent images with different CNN and VT methods, where it aims to discover the adaptability of 
different deep learning models in this research domain.

\begin{figure*}
	\centering
		\includegraphics[scale=.65]{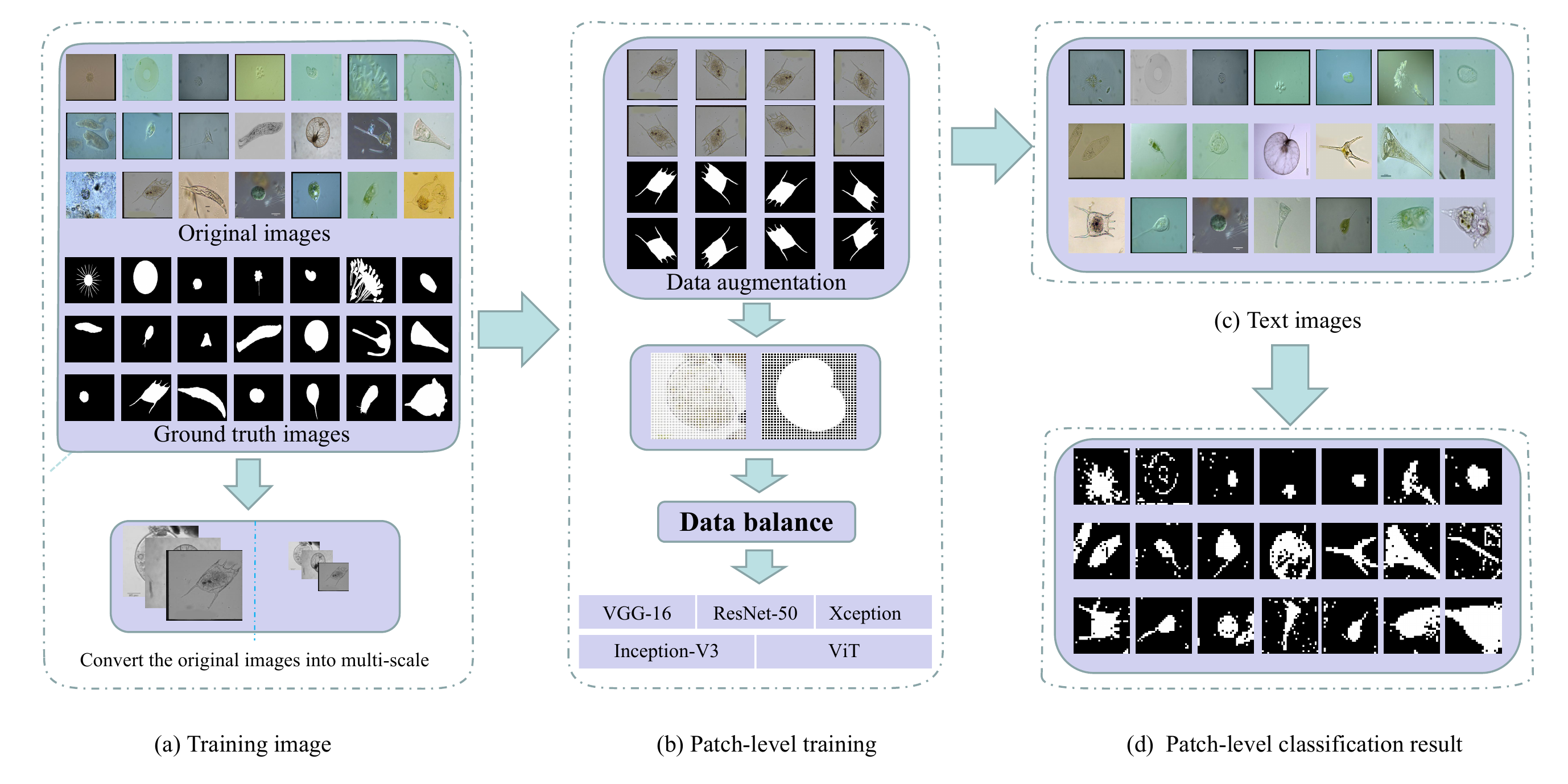}
	\caption{\centering{Workflow of patch-level classification in T-EM images (using environmental microorganism EMDS-5 images as examples).}}\label{Fig:workflow}
\end{figure*}

This paper uses EMDS-5 as an example of T-EM images. 
First, the transparent images are divided into training, 
validation, and test sets according to a ratio of 2:2:4. 
The workflow of patch-level image classification is shown in 
Fig.~\ref{Fig:workflow}, where (a) is the training set, 
including original images and ground truth (GT) images with 
multi-scale settings. (b) is the training process of deep 
learning models where several typical deep learning methods 
are selected and trained. (c) is the test set. 
(d) is the patch-level classification prediction result.

The structure of this paper is as follows: 
In Section~\ref{Sec:related}, related work about deep 
learning in the classification of T-EM images is 
introduced. In Section~\ref{Sec:experiment}, comparative  
experiments about T-EM image classification on 
multi-scale patches with deep learning methods are carried out.    
 In Section~\ref{Sec:conclusion}, the conclusion and future work about T-EM images are summarized in the patch-level 
classification of deep learning methods results and the future development of deep learning methods for analyzing T-EM images.

\section{Related Work}\label{Sec:related}
First, this section introduces the application scenarios of 
EM. Next, we introduce some common 
analysis methods, application scenarios, and research purposes 
of transparent object images. 
Finally, the advantages and disadvantages of some popular deep learning methods are also discussed.

\subsection{Introduction to Environment Microorganism}
The study of EM is essential in the progress of human society and the development of civilization. EM are involved in most areas of humans. 
In the industrial field, the treatment of industrial wastewater is an important issue to protect the environment. Due to the presence of carbohydrates, fats, proteins and other organic matter in wastewater,  which are food for EM. EMs can convert organic matter into less complex compounds and neutralize toxic substances\cite{Madakka-2019-DITTO}. Therefore, EMs effectively degrading harmful substances in industrial wastewater. 
In the medical field, some EMs with pathogens can bring diseases to humans and cause death. However, people also use these microorganisms to create many seedlings to prevent these diseases, so that the human body has resistance to certain diseases, which has played a key role in the development of medicine\cite{Bambini-2009-tuogi}. 
In the agricultural field, the metabolites of some EMs can be used as natural biological pesticides. People use its characteristics to produce agricultural products including pesticides, herbicides, fertilizers and so on. In addition, the \emph{Paramecium} in the rice fields can purify sewage to improve the survival rate of crops and increase production\cite{chapple-2000-TAPOM}. In the ecological field, \emph{rotifers} can decompose the garbage in the water and reduce the degree of eutrophication, thereby improving the quality of water. Moreover, \emph{rotifers} are good biological food for fry growth, which play an important role in balancing the aquatic ecosystem\cite{xu-2020-aefog}. 
In the food field, the \emph{yeast} can be used to make puffed food, edible vinegar, and wine making\cite{dufosse-2006-mpofg}.

\subsection{Introduction to Transparent Object Image Analysis}
Object analysis is one of the essential branches in robot vision, especially the analysis of transparent images  
of objects (transparent object images) is challenging~\cite{zeng-2017-MSDLF}. 
In traditional machine analysis methods, the flexibility of 
transparent object image features obtained by integrating multi-class 
algorithms is poor, and  performance is difficult to improve.
For example, home robots can not see things at all when they are 
detecting some transparent glassware. 
The ClearGrasp machine learning algorithm performs well in  
analyzing transparent objects~\cite{sajjan-2020-CGSEO}. 
It can 
estimate high-precision 
data of transparent objects from RGB-D transparent images, 
thereby improving the accuracy of transparent object detection.

As a necessary technical means for analyzing objects, 
photoelectric sensors are widely used in  
industrial automation, mechanization, and intelligence. 
It uses the properties of light to detect the position 
and change of the object, but when detecting transparent 
color objects, the light beam of the traditional diffuse 
reflection photoelectric sensor penetrates the 
transparent material, causing the sensor to fail. 
Diffuse reflection photoelectric sensor adopts a 
phase-locked loop narrowband filter frequency selection 
technology, which improves the sensitivity to 
self-returning light and stability of detecting transparent 
objects~\cite{chunjiao-2013-TAADO}.

There are many transparent objects in the industrial field, 
such as transparent plastics, transparent colloids, and liquid drops. 
These transparent objects bring much uncertainty to 
products. If factories want to have high-quality products, 
sometimes it is essential to analyze these transparent 
objects and control the shapes of the transparent objects. 
However, it is a difficult problem to segmentation the shape of 
transparent objects through morphological methods. 
For instance, Hata et al. used a genetic algorithm to segmentation 
the transparent paste drop shape in the industry and obtained good 
performance~\cite{hata-1996-SEOTO}.

The segmentation of transparent objects is very useful in 
computer vision applications.
However, the foreground of a transparent image is 
usually similar to 
its background environment, which leads to the general image 
segmentation methods in dealing with transparent images in 
general. 
The light field image segmentation method can accurately and 
automatically segment transparent object images with a small depth 
of field difference and improve the accuracy of the
segmentation, and it has a small amount of 
calculation~\cite{xu-2015-TTOSF}. Hence, it is widely used in the 
segmentation of transparent object images.

The correct segmentation of zebrafish in biology has extensively 
promoted the development of life sciences. 
However, the zebrafish's transparency makes the edges 
blurred in the segmentation. 
The mean shift algorithm can enhance the color representation 
in the image and improve the discrimination of the specimen 
against the background~\cite{guo-2018-AEARH}. 
This method improves the efficiency 
and accuracy of zebrafish specimen segmentation.

Visual object classification is vital for robotics 
and computer vision applications.
Commonly used statistical classification methods such as 
bag-of-features~\cite{nasirahmadi-2017-BOFMF} are 
often applied to image 
classification. The principle is to extract local features 
of the image for classification. 
However, these methods cannot be applied to the 
classification of transparent images because transparent object  
images largely depend on the background. Foreground 
transparent objects do not have their complete 
characteristics, and  it is not easy to classify them accurately. 
The more popular method is the light field distortion 
feature~\cite{xu-2015-LFDFF}, which can describe transparent objects without 
knowing the texture of the scene, thus improving the accuracy 
of classifying transparent object images.

\subsection{Deep Learning}
Simonyan et al. propose the VGG series of deep learning 
network models (VGG-Net), of which VGG-16 is the most 
representative~\cite{simonyan-2014-VDCNF}. 
VGG-Net can imitate a larger receptive field by using multiple 
3×3 filters, enhancing nonlinear mapping, reducing parameters, and improving the network to be more judgmental. 
Meanwhile, VGG-16 continues to deepen the previous 
VGG-Net, with 13 convolutional layers and three fully 
connected layers. 
With the continuous increase of convolution kernel and 
convolution layer, the nonlinear ability of the model is 
stronger. VGG-16 can better learn the features in images 
and achieve good performance in analyzing image 
classification, segmentation, and detection.  
Simonyan proves that as the depth of the network increases, 
it promotes the accuracy of image analysis~\cite{simonyan-2014-VDCNF}. Nevertheless, this increase 
in depth is not without a limit. 
Excessively increasing the depth of the network will lead to 
network degradation problems. Therefore, the optimal network 
depth of VGG-Net is set to 16-19 layers. Moreover, VGG-16 has 
three fully connected layers, which causes more memory to 
be occupied, too long training time, and difficulty in tuning 
parameters.

He et al. propose the ResNet series of networks and add a 
residual structure in networks to solve the problem of network 
degradation~\cite{he-2016-DRLFI}.
The ResNet model introduces a jumpy connection method 
"shortcut connection". This connection method allows the 
residual structure to skip some levels that not be 
fully trained in the feature extraction process and 
increases the model's utilization of feature information 
during the training process.
As the most classical model in the ResNet series, 
ResNet50 has a 50-layer network structure. This model adopts 
the highway network structure, which makes the network have 
strong expression capabilities and the ability to acquire more 
advanced features. 
Therefore, it is widely used in the field of image analysis. 
However, the network model is too deep and complicated, 
so how to judge which layers in the deep network not be fully trained and then optimize the network is 
a complex problem.

Szegedy et al. propose the GoogLeNet network model, 
which has the advantage of reducing the complexity of the 
network based on ResNet. They first proposed 
Inception-v1, whose network is 22 layers deep and consists 
of multiple Inception structures cascade as basic modules.
Each Inception module consists of a 1×1, 3×3, 5×5 convolution 
kernel and a 3×3 maximum pooling, which is similar to the 
idea of multi-scale and increases the adaptability of the 
network to different scales
~\cite{szegedy-2015-GDWCN}. 
With the continuous improvement of the inception module, 
the inception-v2 network uses two 3×3 convolutions instead 
of 5×5 convolutions and increases the BN method, which 
reduces the amount of calculation and speeds up the 
training time~\cite{ioffe-2015-BNADN}. 
The Inception-v3 network introduces the idea of decomposing 
convolution, splitting a larger two-dimensional convolution 
into two smaller one-dimensional convolutions, further 
reducing the amount of calculation~\cite{szegedy-2016-RTIAF}. 
At the same time, Inception-v3 optimizes the Inception 
module, embeds the branch in the branch and improves the 
model's accuracy.

Xception is another improvement after 
Inception-v3~\cite{chollet-2017-XDLWD}. 
It mainly uses depthwise separable convolution to replace 
the convolution operation in Inception-v3. 
The Xception model uses deep separable convolution to 
increase the width of the network, which not only improves  
the accuracy of classification but also improves the 
ability to learn subtle features. 
Meanwhile, Xception adds a residual mechanism 
similar to ResNet to significantly improve the speed of 
convergence during training and the model's accuracy. 
However, Xception is relatively fragmented in the calculation 
process, which results in a slower iteration speed during 
training.

Transformer is a deep neural network based on the 
self-attention mechanism, which enables the model to be 
trained in parallel and obtain the training data's global information. Due to its computational efficiency and 
scalability, it is widely used in  
Natural Language Processing. 
Recently, Dosovitskiy et al. proposed the Vision Transformer 
(ViT) model and found that it performs very well on image 
classification tasks~\cite{dosovitskiy-2020-AIIWW}. 
In the first step of training, the ViT model divides  
pictures into fixed-size image patches and uses its linear
sequence as the input of the transformer model. 
In the second step, position embeddings are added to the 
embeddings patches to retain the position information, and 
then the image features are extracted through the multi-head 
attention mechanism. Finally, the classification model is 
trained. 
ViT breaks through the limitation that RNN model cannot be 
calculated in parallel, and self-attention can produce a 
more interpretable model. 
ViT is suitable for solving image processing tasks, 
but experiments have proved that large data samples are 
needed to improve the training effect.
\subsection{Summary}
T-EM image analysis is used in various fields, 
but the foreground and background of EM transparent images are 
too similar to make analysis difficult. Compared with deep learning methods, the general traditional analysis methods 
are time-consuming, labor-intensive, and costly. 
So this paper compares the performance of several classical 
deep learning networks for T-EM image analysis.
\section{Comparative Experiment}\label{Sec:experiment}
This section introduces the patch-level classification experiment 
and classification results of T-EM images under several deep learning networks.
\subsection{Experiment Setting}
\subsubsection{Data Settings} 
\begin{figure*}
	\centering
		\includegraphics[scale=.70]{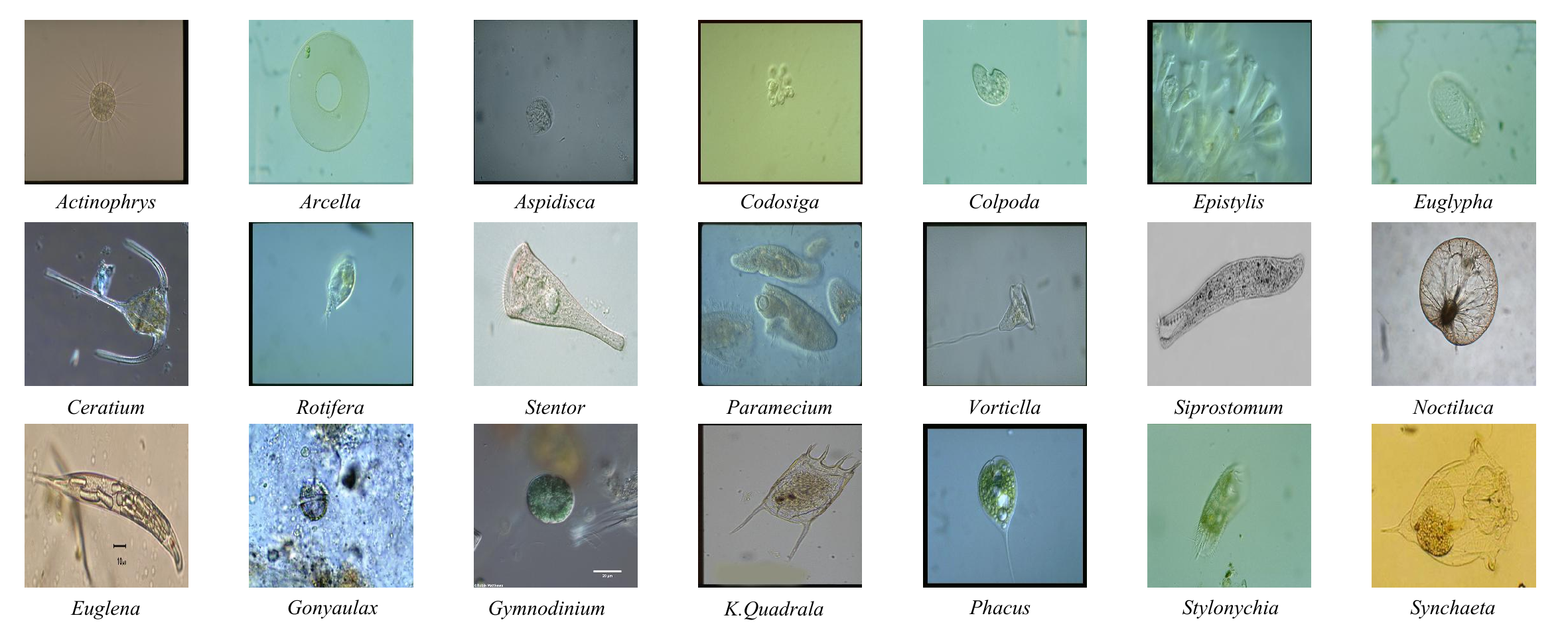}
	\caption{\centering{Examples of the environmental microorganism image in EMDS-5.}}\label{Fig:EMDS5}
\end{figure*}

\begin{figure*}
	\centering
		\includegraphics[scale=.70]{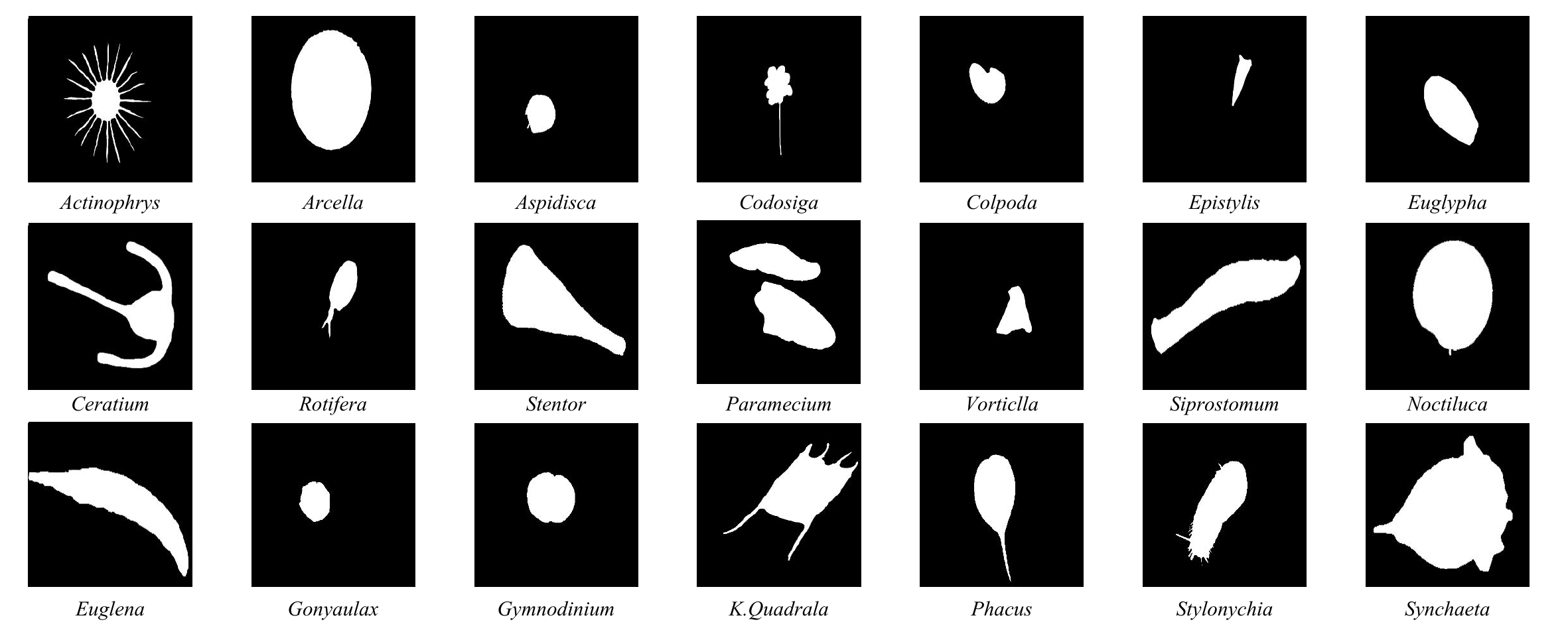}
	\caption{\centering{Environmental microorganism EMDS-5 ground truth images.}}\label{Fig:EMDS5-GT}
\end{figure*}

In our work, we use EM Data Set Fifth Version (EMDS-5) as T-EM images for analysis~\cite{li-2021-EMIDF}. 
It is a newly released version of the EMDS 
series, which contains 21 types of EM, each of which contains 
20 original microscopic images and their corresponding ground 
truth (GT) images (examples are shown in Fig.\ref{Fig:EMDS5} and Fig.~\ref{Fig:EMDS5-GT}).
We randomly divide each category of EMDS-5 into 
training, validation, and test data sets at a ratio of 
1:1:2. Therefore, as shown in Tab.1, we have 105 original images and their 
corresponding GT images for training and validation, 
respectively, and 210 original images for testing.

\subsubsection{Data Preprocessing}

\begin{table}
\caption{EMDS-5 Experimental data.}\label{tbl1}
\begin{tabular*}{\tblwidth}{@{} LLLL@{} }
\toprule
            & Training Set & Validation Set & Test Set \\
\midrule
\emph{Actinophrys} &\makecell[c]{5}&\makecell[c]{5}&\makecell[c]{10}\\
\emph{Arcella}    &\makecell[c]{5}&\makecell[c]{5}&\makecell[c]{10}\\
\emph{Aspidisca}   &\makecell[c]{5}&\makecell[c]{5}&\makecell[c]{10}\\
\emph{Codosiga}    &\makecell[c]{5}&\makecell[c]{5}&\makecell[c]{10}\\
\emph{Colpoda}     &\makecell[c]{5}&\makecell[c]{5}&\makecell[c]{10}\\
\emph{Epistylis}  &\makecell[c]{5}&\makecell[c]{5}&\makecell[c]{10}\\
\emph{Euglypha}    &\makecell[c]{5}&\makecell[c]{5}&\makecell[c]{10}\\
\emph{Paramecium}  &\makecell[c]{5}&\makecell[c]{5}&\makecell[c]{10}\\
\emph{Rotifera}    &\makecell[c]{5}&\makecell[c]{5}&\makecell[c]{10}\\
\emph{Vorticlla}   &\makecell[c]{5}&\makecell[c]{5}&\makecell[c]{10}\\
\emph{Noctiluca}   &\makecell[c]{5}&\makecell[c]{5}&\makecell[c]{10}\\
\emph{Ceratium}    &\makecell[c]{5}&\makecell[c]{5}&\makecell[c]{10}\\
\emph{Stentor}     &\makecell[c]{5}&\makecell[c]{5}&\makecell[c]{10}\\
\emph{Siprostomum} &\makecell[c]{5}&\makecell[c]{5}&\makecell[c]{10}\\
\emph{K.Quadrala}  &\makecell[c]{5}&\makecell[c]{5}&\makecell[c]{10}\\
\emph{Euglena}     &\makecell[c]{5}&\makecell[c]{5}&\makecell[c]{10}\\
\emph{Gymnodinium} &\makecell[c]{5}&\makecell[c]{5}&\makecell[c]{10}\\
\emph{Gonyaulax}   &\makecell[c]{5}&\makecell[c]{5}&\makecell[c]{10}\\
\emph{Phacus}      &\makecell[c]{5}&\makecell[c]{5}&\makecell[c]{10}\\
\emph{Stylonychia} &\makecell[c]{5}&\makecell[c]{5}&\makecell[c]{10}\\
\emph{Synchaeta}   &\makecell[c]{5}&\makecell[c]{5}&\makecell[c]{10}\\
total       &\makecell[c]{105}&\makecell[c]{105}&\makecell[c]{210}\\
\bottomrule
\end{tabular*}
\end{table}

In the first step, we uniformly convert all images sizes to 
$224 \times 224$ pixels and 7168×7168 pixels to keep that each image is cropped
into the same number of multi-scale patches. 
In the second step, we gray-scale EMDS-5 images to 
facilitate the calculation of gradients and feature 
extraction during training. 
In the third step, the training and validation images, 
and their corresponding GT images are 
cropped into patches ($8 \times 8$ pixels and $224 \times 224$ pixels),  
where $105 \times 1024$=107520 patches are obtained. 
We divide these small patches into two categories according 
to the corresponding GT image small patches: foreground 
and background. The partition criterion is whether the target area in 
the patch takes up half of the whole patch. If it is, we will assign 
foreground as the label of this patch; otherwise, it will be 
background. 
In the fourth step, we find that the $224 \times 224$ pixel patches with 
foreground and background are 16630 and 90890, respectively. 
In order to avoid data imbalance during training, 
we rotate the training set image small patches by 0, 90, 
180, 270 degrees and mirror them for data augmentation. 
Then we further obtain 16630×8=133040 patches, from which 
90890 patches 
are randomly selected as the finally used patches in the training 
set. 
We expand the data of the $8 \times 8$ pixel patches according to the same process.
The details of the augmented data are provided in Tab.~\ref{tbl2}.

\begin{table}
\caption{Data augmentation. FG (foreground) and BG (background)}\label{tbl2}
\renewcommand\arraystretch{1.8}
\begin{tabular*}{\tblwidth}{@{} LLL@{} }
\toprule
\makecell[c]{Data Set}&\makecell[c]{Training Set}&\makecell[c]{Validation Set}\\
\midrule
\makecell[c]{$8 \times 8$ pixel FG} & \makecell[c]{16554} & \makecell[c]{17356}\\
\makecell[c]{$8 \times 8$ pixel BG} & \makecell[c]{90966} & \makecell[c]{90164}\\
\makecell[c]{Augmentation With FG} & \makecell[c]{90966} & \makecell[c]{$\backslash$} \\
\makecell[c]{$8 \times 8$ Total} & \makecell[c]{181932} & \makecell[c]{107520}\\
\midrule
\makecell[c]{$224 \times 224$ pixel FG} & \makecell[c]{16630} & \makecell[c]{17459}\\
\makecell[c]{$224 \times 224$ pixel BG} & \makecell[c]{90890} & \makecell[c]{90061}\\
\makecell[c]{Augmentation With FG} & \makecell[c]{90890} & \makecell[c]{$\backslash$} \\
\makecell[c]{$224 \times 224$ Total} & \makecell[c]{181780} & \makecell[c]{107520}\\
\bottomrule
\end{tabular*}
\end{table}

\subsubsection{Experimental Environment}
Our classification comparison experiment is conducted on a local 
computer with Win10 Professional operating system, and the computer 
has 16 GB RAM i7-10700 CPU and 8 GB NVIDIA Quadro RTX 4000 GPU. 
The CNN models we use in this paper are based on the Keras 2.3.1 
framework using Tensorflow 2.0.0 as the backend; in the ViT model, 
we use the Pytorch 1.7.1 and Torchvision 8.0.2 operating environment.

\subsubsection{Hyper Parameters}
\begin{figure*}
	\centering
		\includegraphics[scale=.5]{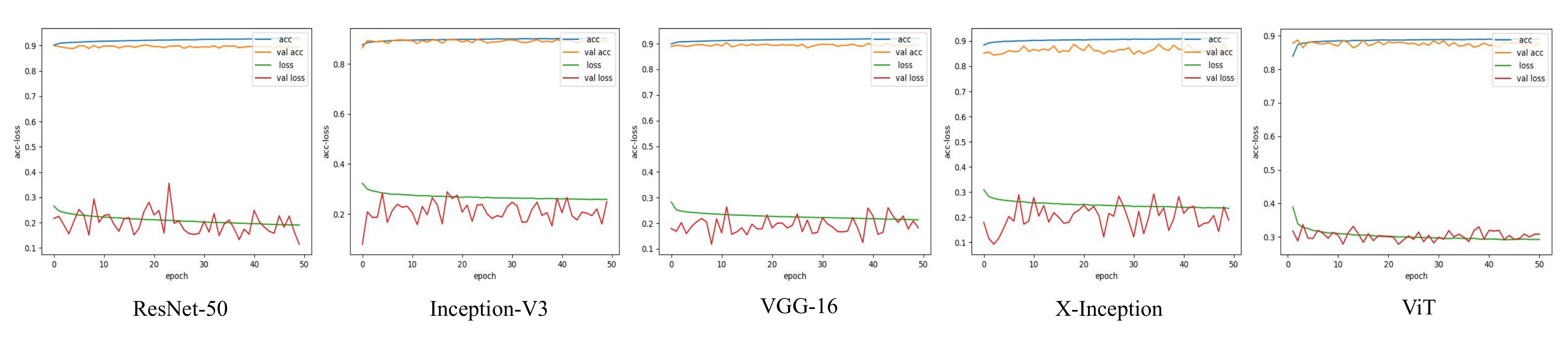}
	\caption{\centering{Compare the results of the loss and accuracy curves of deep learning on the $8 \times 8$ pixels training and the validation sets.}}\label{Fig:model-8}
\end{figure*}

\begin{figure*}
	\centering
		\includegraphics[scale=.5]{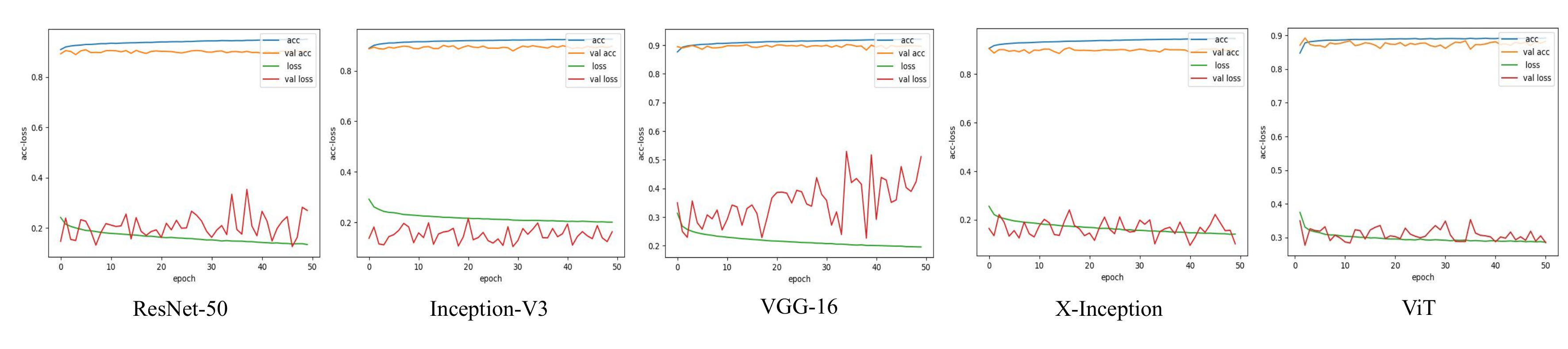}
	\caption{\centering{Compare the results of the loss and accuracy curves of deep learning on the $224 \times 224$ pixels training and the validation sets.}}\label{Fig:model-224}
\end{figure*}

This experiment uses the Adam optimizer with a 0.0002 learning rate 
and sets the batch size to 32 in our training process. In Fig.~
\ref{Fig:model-8} and Fig.~\ref{Fig:model-224} 
show the accuracy and loss curves of different deep learning models 
in this experiment. 
We find that the loss and accuracy curves of 
the training set are converging after training for 40 layers.  
Therefore, considering the computational 
performance of the workstation, we finally set 50 epochs for training.

\subsection{Evaluation Metrics} 
To compare the classification performance of different methods, 
we used the commonly used deep learning classification indexes 
Accuracy (Acc), Precision (Pre), Recall (Rec), Specificity (Spe), and F1-Score (F1) to evaluate 
the classification results. Acc reflects the ratio of correct 
classification samples to total samples. Pre reflects 
the proportion of correctly predict positive samples in the positive 
samples of model classification. Rec reflects the correct proportion 
of model classification in total positive samples Spe 
reflects the proportion of the model correctly classifying the 
negative samples in the total negative samples. F1 is a calculation result that comprehensively considers the Pre and Rec of the model. 
These evaluation indexes are defined in Tab.~\ref{tbl3}. TP (True Positive), FN (False Negative), FP (False Positive), and TN (True Negative) are concepts in the confusion matrix.

\begin{table}
\caption{Evaluation metrics for images classification.}\label{tbl3}
\renewcommand\arraystretch{1.8}
\begin{tabular*}{\tblwidth}{@{} LL@{} }
\toprule
\makecell[c]{Assessments} & \makecell[c]{Formula} \\
\midrule
\makecell[c]{Acc} & \makecell[c]{$\rm \frac{TP+TN}{TP + TN + FP + FN}$} \\
\makecell[c]{Pre} & \makecell[c]{$\rm \frac{TP}{TP + FP}$} \\
\makecell[c]{Rec} & \makecell[c]{$\rm \frac{TP}{TP + FN}$} \\
\makecell[c]{Spe} & \makecell[c]{$\rm \frac{TN}{TN + FP}$} \\
\makecell[c]{F1} & \makecell[c]{$ 2 \times \frac{P \times R}{P + R}$} \\
\bottomrule
\end{tabular*}
\end{table}

\subsection{Comparative Experiment}
\subsubsection{Comparative Experiment of $8 \times 8$ Pixel Patches}
\paragraph{\textbf{Comparison on Training and Validation 
Sets:}} In order to compare the classification performance of 
CNNs and ViT models, we calculate Pre, Rec, 
Spe, F1, Acc, Max Acc.  
In Tab.~\ref{tbl4}, we summarize the results of $8 \times 8$ pixel 
patches on the validation set for each model.
Overall, the Pre of the deep learning network 
in classifying the T-EM image backgrounds is 
higher than the foreground. 
Besides, the Pre of the five models to classify T-EM image backgrounds is almost 97$\%$, the highest is the VGG-16 
value of 97.6$\%$, and the lowest is the X-Inception and the ViT 
value of 96.7$\%$. 
Meanwhile, the Pre rate of classification foreground VGG-16 
is the best and the Pre rate is 63.1$\%$. The 
Inception-V3 obtains the lowest 53.3$\%$. For T-EM image foregrounds classification, the highest Rec rate is obtained with X-Inception, which is 89.2$\%$, and the lowest one is ViT, which is 84.1$\%$.
For T-EM image backgrounds classification, the 
highest Rec rate is the Vit value of 90.3$\%$ and the 
lowest is the X-Inception value of 85.0$\%$. The spe  
obtained by the five models in the classification 
background is opposite to the Rec rate obtained in the 
classification foreground. 
Among the five models, the highest Acc is ResNet50 with a 
value of 92.87$\%$, and the lowest is ViT with a value of 
89.26$\%$.

\begin{table}\small
\caption{A comparison of the classification results on validation set of $8 \times 8$ pixel patches.
MAcc (Max Acc), FG (foreground) and BG (background) (In [\%].)}
\label{tbl4}
\begin{tabular}{@{}llllllll@{}}
\toprule
\multicolumn{1}{l}{Model}      & Class    & Pre & Rec  & Spe & F1 & \multicolumn{1}{l}{MAcc} \\
\midrule
\multirow{2}{*}{ResNet50}       & FG    & 62.3     & 88.2  & 89.7      & 73.0   & \multirow{2}{*}{92.87}                            \\
                                & BG    & 97.5     & 89.7   & 88.2     & 93.4                                                               \\
\multirow{2}{*}{Inception-V3}   & FG    & 61.8     & 88.6  & 89.5      & 72.8   & \multirow{2}{*}{90.24}                 \\
                                & BG    & 97.6     & 89.5  & 88.6      & 93.4                                                                 \\
\multirow{2}{*}{VGG-16}         & FG    & 63.1     & 88.6  & 90.0      & 73.7    & \multirow{2}{*}{92.09}                \\
                                & BG    & 97.6     & 90.0  & 88.6      & 93.6     \\
\multirow{2}{*}{X-Inception}    & FG    & 53.3     & 89.2  & 85.0      & 66.7    & \multirow{2}{*}{91.10}                 \\
                           & BG & 96.7     & 85.0  & 89.2      & 90.9      \\           
\multirow{2}{*}{ViT}            & FG    & 62.4     & 84.1  & 90.3      & 71.6   & \multirow{2}{*}{89.26}               \\
                           & BG & 96.7     & 90.3  & 84.1      & 93.4   \\           
\bottomrule
\end{tabular}
\end{table}

\paragraph{\textbf{Comparison on Test Set:}} In Tab.~\ref{tbl5} we 
summarize the results of these five network predictions. we can find 
that Acc of ResNet50 is the highest (90.00$\%$), 
Acc of X-Inception is the lowest at 85.85$\%$. Furthermore, 
the lowest prediction Acc of 
the transparent foreground is the X-Inception value of 51.8$\%$, and 
the highest is the ResNet50 value of 62.2$\%$.

\begin{figure*}
	\centering
		\includegraphics[scale=.7]{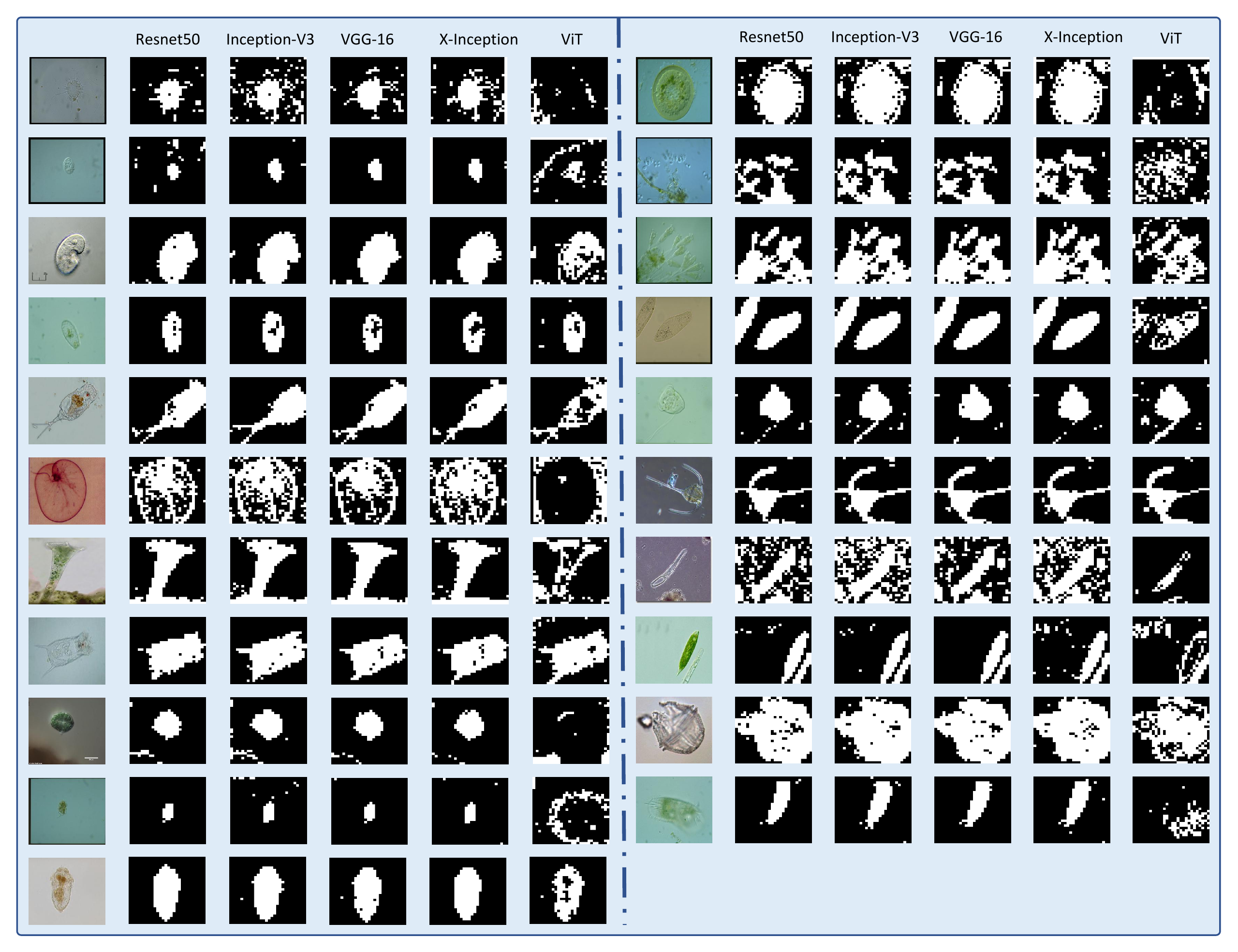}
	\caption{\centering{Reconstruction of $8 \times 8$ pixel T-EM images 
	classification results.}}\label{Fig:result}
\end{figure*}

\begin{figure*}
	\centering
		\includegraphics[scale=.5]{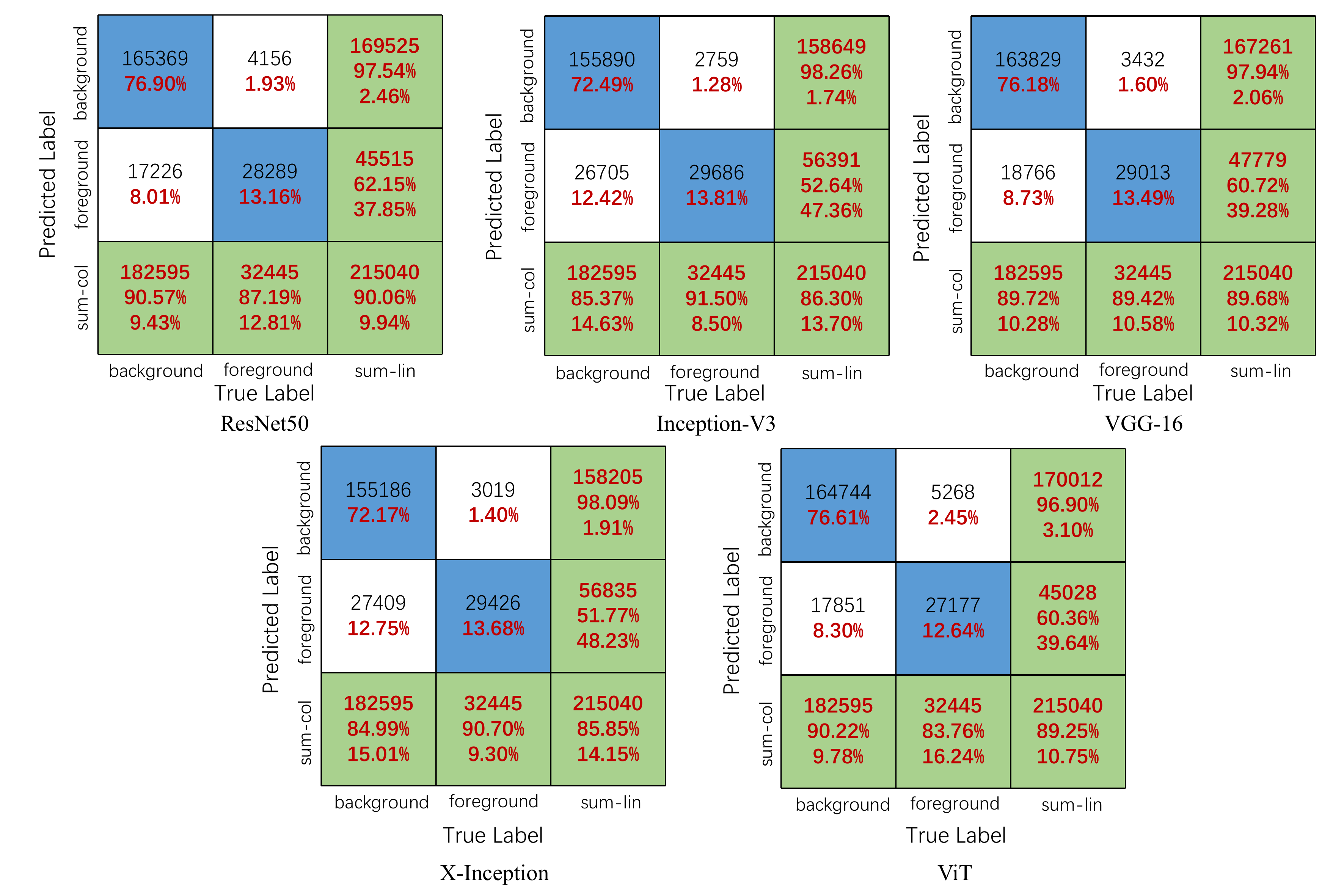}
	\caption{\centering{Predict the confusion matrix on test set of $8 \times 8$ pixel patches}}\label{Fig:Confusion matrix}
\end{figure*}

In order to more intuitively express the classification results of CNN 
and ViT models for T-EM image patches, we summarize the 
confusion matrices predicted by five models into Fig.~\ref{Fig:Confusion matrix}. 
We find that the ability of CNNs to classify foreground 
patches of T-EM images is higher than that of ViT. Among 
them, the best CNN models is Inception-V3, which correctly 
classify 29686 foreground patches, accounting for 91.50$\%$ 
of the total correct foreground patches. ViT correctly 
classify 27177 foreground patches, accounting for 83.76$\%$ 
of the total correct foreground patches. In addition, the 
number of correctly classify backgrounds in ResNet50 is at 
most 165369, accounting for 90.57$\%$ of the total correct 
background patches, and the Pre of the classify background 
patches is 97.55$\%$. Among the five models, ResNet50 has the 
highest prediction accuracy rate of 90.06$\%$ 
To better show the classification results, we reconstruct 
T-EM images after dicing in Fig.~\ref{Fig:result}.

In Tab.~\ref{tbl4.1}, we provide the model training and prediction time and the size of the model during the experiment.
From the perspective of model training time, the ViT 
model is much lower than CNN models, where the ViT 
training time is 12418 seconds, and the X-Inception 
training time is the longest 45897 seconds. 
From the perspective of the size of the model, 
the minimum size of the ViT model is 31.2M, 
and the maximum size of the ResNet50 model is 114M.  
We calculate the time of the five prediction models. The fastest 
prediction time of VGG-16 is 757 seconds and the prediction time of a 
single picture is 0.0063 second. The slowest time of ViT is 1670 
seconds and the prediction time of a single picture is 0.0078 second.

\begin{table}
\caption{A comparison of the classification results on train and test sets of $8 \times 8$ pixel patches. Train (Train times), Test (Test times) and Avg (Single picture prediction time )(In [s].)}
\label{tbl4.1}
\begin{tabular}{@{}llllll@{}}
\toprule
  model        & Train & Test & Avg & Size(MB) \\
\midrule
ResNet50     & 48762   & 1448       & 0.0067    & 114      \\
Inception-V3 & 61443   & 1186       & 0.0055    & 107      \\
VGG-16       & 49477   & 757        & 0.0035    & 62.2     \\
X-inception  & 61247   & 999        & 0.0046    & 103      \\
ViT          & 22133   & 1670       & 0.0078    & 31.2     \\
\bottomrule 
\end{tabular}
\end{table}

\subsubsection{Comparative Experiment of $224 \times 224$ Pixel Patches}
\paragraph{\textbf{Comparison on Training and Validation 
Sets:}}We compare the $224 \times 224$ pixel patches in the same way as in the 
previous work in Section 3.1.3. In 
Tab.~\ref{tbl6}, we summarize the classification results 
on validation set of $224 \times 224$ pixel patches for 
each network model. The highest Pre of classification 
foreground is ResNet50, which is 64.6$\%$, the lowest is 
X-Inception value of 60.8$\%$. However, the highest Pre of 
background classification is X-Inception value of 97.9$\%$, and the lowest is ViT value of 96.0$\%$.
The highest Rec of the five model classification 
foreground is that the X-Inception, which is 90.4$\%$, and 
the lowest is the ViT value of 80.7$\%$. 
The spe rate obtained by the five models in the classification background is opposite to the Rec rate obtained in the classification foreground.
Overall, the Rec and Spe rate of the five model 
classification backgrounds are almost 90$\%$.
When ResNet50 classifies the foreground and background of T-EM images, F1-Score achieves the best result. 
Meanwhile, the Acc of ResNet50 model training is the 
highest at 94.99$\%$.

\begin{table}
\caption{A comparison of the classification results on test set of $8 \times 8$ pixel patches. PAcc (prediction  accuracy), FG (foreground) and BG (background)(In [\%].)}
\label{tbl5}
\begin{tabular}{@{}lllllllll@{}}
\toprule
\multicolumn{1}{l}{Model}      & Class    & Pre & Rec  & Spe & F1 & \multicolumn{1}{l}{PAcc} \\
\midrule
\multirow{2}{*}{ResNet50}      & FG & 62.2     & 87.2  & 90.6      & 72.6   & \multirow{2}{*}{90.0}                \\
                               & BG & 97.5     & 90.6  & 87.2      & 93.9   \\
\multirow{2}{*}{Inception-V3}  & FG & 52.6     & 91.5  & 85.4      & 66.8   & \multirow{2}{*}{86.29}                \\
                               & BG & 98.3     & 85.4  & 91.5      & 91.4   \\
\multirow{2}{*}{VGG-16}        & FG & 60.7     & 89.4  & 89.7      & 72.6   & \multirow{2}{*}{89.6}                  \\
                               & BG & 97.9     & 89.7  & 89.4      & 93.6    \\
\multirow{2}{*}{X-Inception}   & FG & 51.8     & 90.7  & 85.0      & 65.9    & \multirow{2}{*}{85.85}                    \\
                               & BG & 98.1     & 85.0  & 90.7      & 91.1     \\           
\multirow{2}{*}{ViT}           & FG & 60.4     & 83.8  & 90.2      & 70.2   & \multirow{2}{*}{89.25}                 \\
                               & BG & 96.9     & 90.2  & 83.8      & 93.4   \\           
\bottomrule
\end{tabular}
\end{table}

\paragraph{\textbf{Comparison on Test set:}}
In Tab.~\ref{tbl7}, we summarize the results of these five 
network predictions. we can find that the prediction Acc 
of X-Inception is the highest (89.11$\%$), and the 
prediction Acc of Inception-V3 is the lowest at 88.10$\%$. 
However, the highest Pre in predicting transparent  
foreground is the ViT which is 60.6$\%$.

\begin{table}
\caption{A comparison of the classification results on validation set of $224 \times 224$ pixel patches. MAcc (Max Acc), FG (foreground) and BG (background)(In [\%].)}
\label{tbl6}
\begin{tabular}{@{}lllllll@{}}
\toprule
\multicolumn{1}{l}{Model}      & Class    & Pre & Rec  & Spe & F1 & \multicolumn{1}{l}{Max Acc} \\
\midrule
\multirow{2}{*}{ResNet50}       & FG    & 64.6     & 87.6  & 90.7 & 74.4   & \multirow{2}{*}{94.99}\\
                                & BG & 97.4     & 90.7   & 87.6   & 93.9     \\
\multirow{2}{*}{Inception-V3}   & FG    & 63.0     & 88.9  & 89.9 & 73.7   & \multirow{2}{*}{92.51}       \\
                                & BG & 97.7     & 89.9  & 88.9    & 93.6                    \\
\multirow{2}{*}{VGG-16}         & FG    & 63.2     & 85.8  & 90.3 & 72.8    & \multirow{2}{*}{92.08}                    \\
                                & BG & 97.1     & 90.3  & 85.8    & 93.6    \\
\multirow{2}{*}{X-Inception}    & FG    & 60.8     & 90.4  & 88.7 & 72.7    & \multirow{2}{*}{94.72}              \\
                                & BG & 97.9     & 88.7  & 90.4    & 93.1     \\           
\multirow{2}{*}{ViT}            & FG    & 63.3     & 80.7  & 90.9 & 70.9   & \multirow{2}{*}{89.28}                \\
                                & BG & 96.0     & 90.9  & 80.7    & 93.4    \\           
\bottomrule
\end{tabular}
\end{table}

\begin{figure*}
	\centering
		\includegraphics[scale=.7]{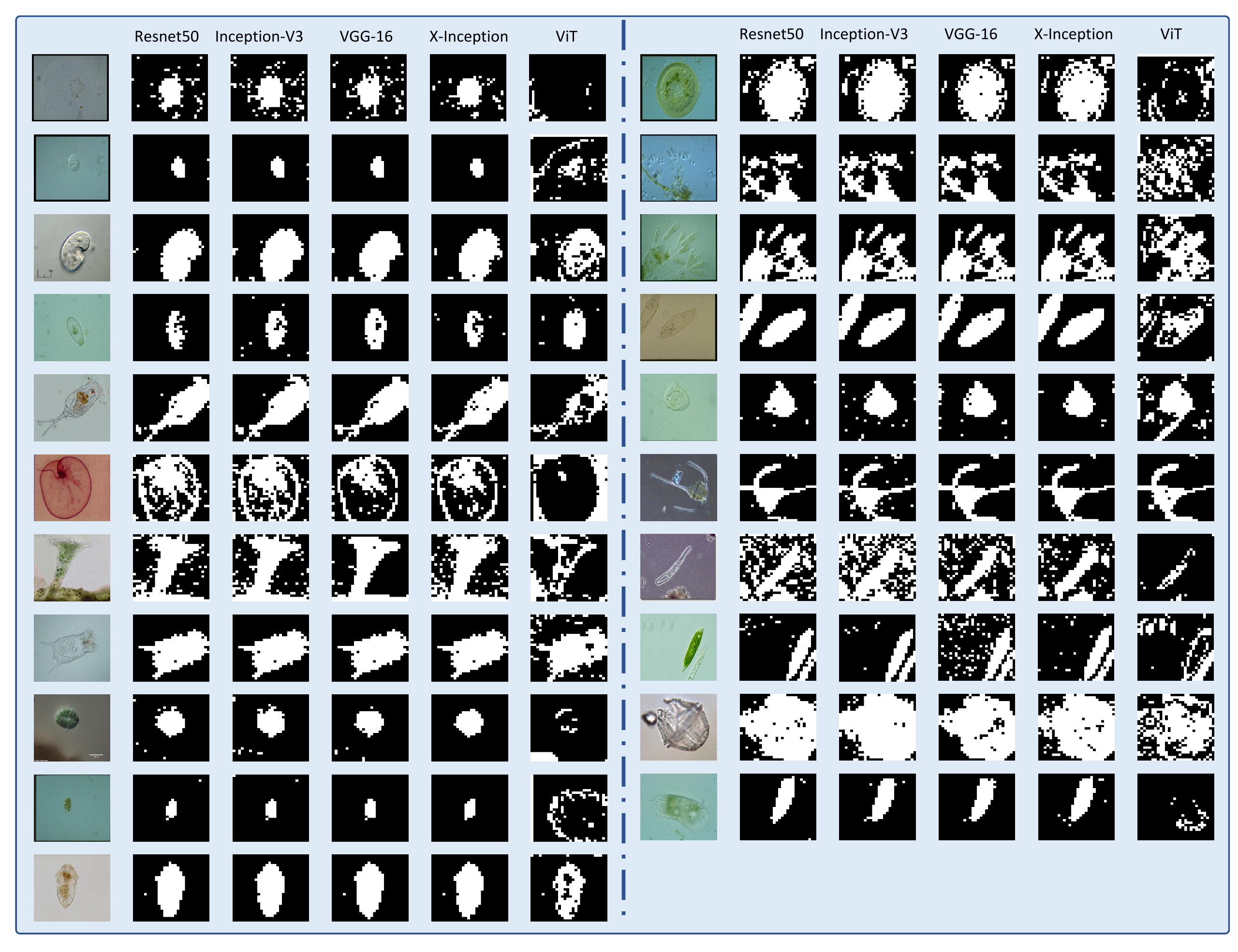}
	\caption{\centering{Reconstruction of $224 \times 224$ pixel T-EM images 
	classification results.}}\label{Fig:result-224}
\end{figure*}

\begin{figure*}
	\centering
		\includegraphics[scale=.5]{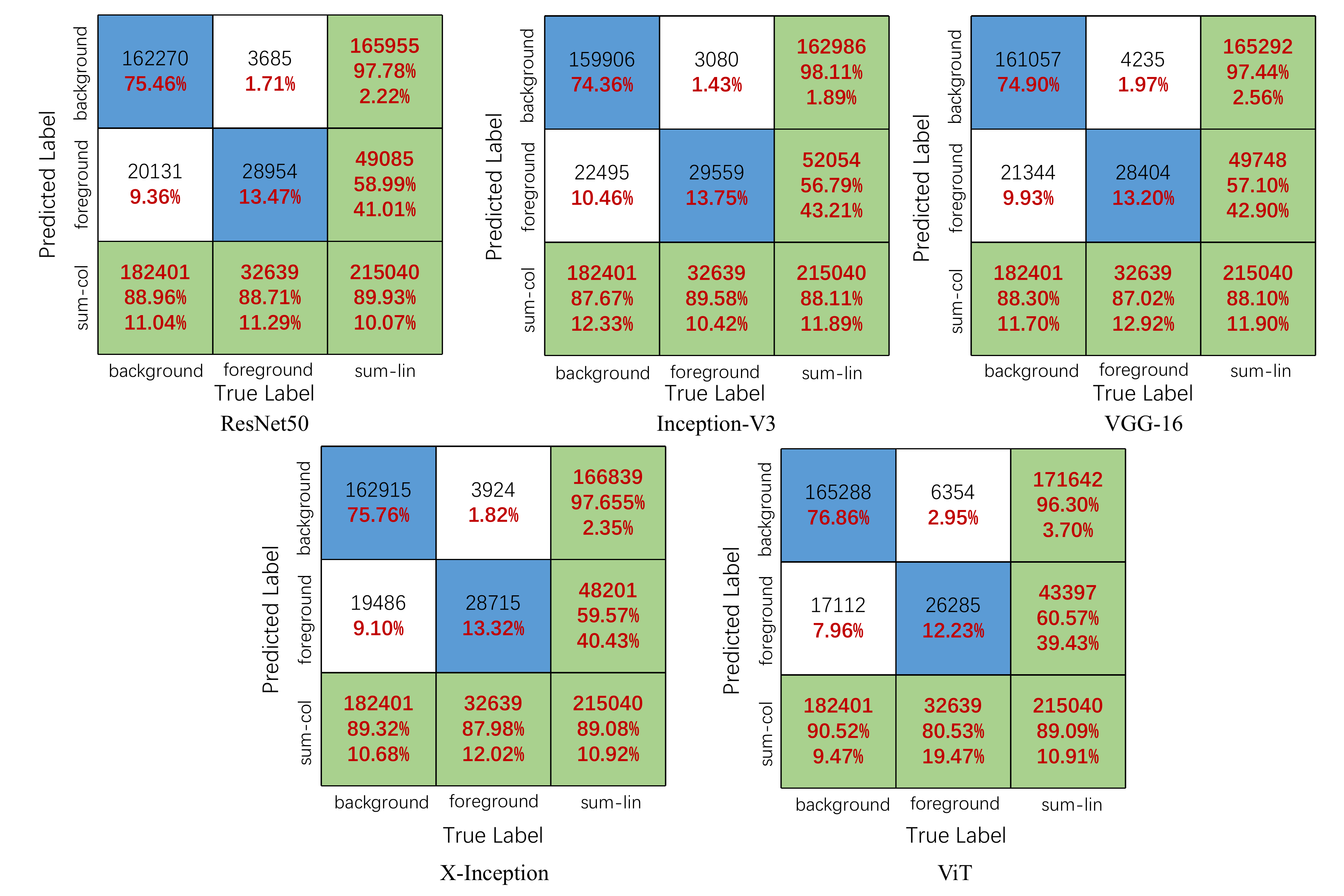}
	\caption{\centering{Predict the confusion matrix on test set of $224 \times 224$ pixel patches}}\label{Fig:Confusion matrix-224}
\end{figure*}

In order to more intuitively express the classification 
results of CNN and ViT models on T-EM image 
patches, we  summarize the confusion matrices predicted by 
five models in Fig.~\ref{Fig:Confusion matrix-224}. We 
find that the ability of CNNs to classify foreground 
patches of T-EM images is higher than that of ViT. 
Among them, X-Inception is the best in the CNN  
models, which correctly classifies 29559 small patches. 
ViT correctly classifies 26285 foreground patches. 
However, the highest accuracy of classifying T-EM image backgrounds is the ViT , which is 90.52$\%$. It 
correctly classifies 165288 background images. Meanwhile, 
in order to better show the classification results, we 
reconstruct T-EM images after dicing in 
Fig.~\ref{Fig:result-224}.

\begin{table}
\caption{A comparison of the classification results on test set of $224 \times 224$ pixel patches. PAcc (prediction  accuracy), FG (foreground) and BG (background)(In [\%].)}
\label{tbl7}
\begin{tabular}{@{}lllllll@{}}
\toprule
\multicolumn{1}{l}{Model}      & Class    & Pre & Rec  & Spe & F1 & \multicolumn{1}{l}{PAcc} \\
\midrule
\multirow{2}{*}{ResNet50}       & FG    & 59.0     & 88.7  & 89.0      & 70.9& \multirow{2}{*}{88.92}\\
                                & BG & 97.8     & 89.0   & 88.7     & 93.2    \\
\multirow{2}{*}{Inception-V3}   & FG    & 56.8     & 90.6  & 87.7      & 69.8   & \multirow{2}{*}{88.10}\\
                                & BG & 98.1     & 87.7  & 90.6      & 92.6   \\
\multirow{2}{*}{VGG-16}         & FG    & 57.1     & 87.0  & 88.3      & 68.9    & \multirow{2}{*}{88.11}\\
                                & BG & 97.4     & 88.3  & 87.0      & 92.6     \\
\multirow{2}{*}{X-Inception}    & FG    & 59.6     & 88.0  & 89.3      & 71.1  & \multirow{2}{*}{89.11}\\
                                & BG & 97.6     & 89.3  & 88.0      & 93.3  \\           
\multirow{2}{*}{ViT}            & FG    & 60.6    & 80.5  & 90.6      & 69.1   & \multirow{2}{*}{89.09}\\
                                & BG & 96.3     & 90.6  & 80.5      & 93.4    \\           
\bottomrule
\end{tabular}
\end{table}

In Tab.~\ref{tbl4.2}, we find that the training time of the ViT model is still the fastest at 12418 seconds, and the slowest is 73465 seconds for X-Inception. Besides the fastest prediction time of Inception-V3 is 1049 seconds and the prediction time of a 
single picture is 0.0060 second. The slowest time of ViT is 2156 
seconds and the prediction time of a single picture is 0.0100 second.

\begin{table}
\caption{A comparison of the classification results on train and test 
sets of $224 \times 224$ pixel patches. Train (Train times), Test (Test 
times) and Avg (Single picture prediction time)(In [s].)}
\label{tbl4.2}
\begin{tabular}{@{}llllll@{}}
\toprule
  model        & Train & Test & Avg & SIZE(MB) \\
\midrule
ResNet50     & 51077   & 1634       & 0.0076    & 114      \\
Inception-V3 & 66095   & 1296       & 0.0060    & 107      \\
VGG-16       & 50908   & 1364        & 0.0063    & 62.2     \\
X-inception  & 73465   & 1049        & 0.0049    & 103      \\
ViT          & 23102   & 2156       & 0.0100    & 31.2     \\
\bottomrule 
\end{tabular}
\end{table}

\subsection{In-depth Analysis}
We compare classification results in Tab.~\ref{tbl4} and Tab.~\ref{tbl6}, and find that when the T-EM image patches of the training input increase from $8 \times 8$ to $224 \times 224$ 
pixels, the Pre of the classification foreground of the five models improves. The biggest improvement is X-Ineption from 53.3$\%$ to 60.8$\%$. 
This shows that when the image input size becomes larger,  
CNNs can extract more features, thereby improving 
the model's classification performance for T-EM images. 
As the input image size increases, it has little effect on the 
performance of the five models to classify the T-EM image 
backgrounds. Among them, the biggest improvement is X-Inception, which has 
an Acc increase of 1.2$\%$. 
With the increase in the size of the input image of the CNNs, the training time of the five models has also increase 
by 2$\%$ to 6$\%$, but the Acc of the model has also improve, and 
the training Acc of the ResNet50 model is the highest value of 
94.99$\%$. VGG-16 and ViT remain basically unchange.

We compare Tab.~\ref{tbl5} and Tab.~\ref{tbl7}. When the T-EM image is cropped into 
pathes of $8 \times 8$ pixels, the prediction Acc of the ResNet50 model 
is the highest value of 90$\%$, and the lowest is the X-Inception value 
of 85.85$\%$. However, when the T-EM image is enlarged and cropped 
into patches of $224 \times 224$ pixels, the X-Inception prediction Acc 
rate is the highest 89.11$\%$, and the second is that the ViT Acc 
rate is 89.09$\%$. By increasing the size of the input T-EM image 
patches, the prediction Acc of the ViT model exceeds that of ResNet50, 
Inception-V3 and VGG-16. Moreover, when the input image is a small patches 
of $224 \times 224$ pixels, the Pre of the ViT model to classify the 
foreground of the T-EM image is higher than that of the CNNs 
network. This shows that the advantage of ViT for global information 
description is higher than that of some CNNs networks.

In the predicted 215040 patches, we compare the performance of 
five types of network classification foreground and background. 
In Fig.~\ref{Fig:Confusion matrix}, 
we find that Inception-v3 has the largest number of correct foregrounds 
under $8 \times 8$ pixel patches. ResNet50 has the largest number of 
correctly classify backgrounds. In Fig.~\ref{Fig:Confusion matrix-224} 
we find that Inception-v3 
has the largest number of correctly classify foregrounds under $224 \times 224$ 
pixel patches, and the largest number of correctly classify 
background  patches is ViT. In addition, the number of foreground 
patches misclassify by the ViT network model is much smaller than that 
of the CNNs network. At the same time, the number of correctly classify 
foreground in the CNNs network is greater than that of the ViT network. 

\section{Conclusion and Future Work}\label{Sec:conclusion}
In this paper, we aim at the problem that T-EM images are difficult to classify by cropping the image 
into patches and classifying the foreground and background. We 
use CNNs (ResNet50, Inception-V3, VGG-16, X-Inception) and ViT 
deep learning methods to compare the performance of classifying 
patches of T-EM images. In addition, we also 
compare the effects of patches of 
$8 \times 8$ and $224 \times 224$ pixels 
on the classification performance of deep learning methods. We 
conclude that CNNs have better classification performance than 
ViT in patches of 8×8 pixels. However, the classification 
performance of ViT at 256×256 pixels is better than that of 
most CNNs. Therefore, we conclude that CNNs and ViT network 
models have more advantages in image classification. CNNs are 
good at extracting local features of images, and ViT is good at extracting images global features.

In the future, we plan to increase the amount 
of data to improve the stability of the 
comparison. Meanwhile, the images reconstructed by deep 
learning classification can be extended to the positioning, 
segmentation, recognition, and detection of T-EM images. We will further strengthen the application of results.

\section{Acknowledgements}
This work is supported by National Natural Science Foundation 
of China (No. 61806047). We thank Miss Zixian Li and Mr. 
Guoxian Li for their important discussion.

\bibliographystyle{unsrt}
\bibliography{wenxian}
\end{document}